\documentclass[twoside]{IEEEtran}
\usepackage{graphics} 
\usepackage[hyphens]{url}
\usepackage{lipsum}
\usepackage{listings}
\usepackage[font=footnotesize]{caption}
\usepackage{subcaption}
\usepackage{epsfig} 
\usepackage{times} 
\usepackage{amsmath} 
\usepackage{amsfonts}
\usepackage{cases}
\usepackage{color}
\usepackage{cases}
\usepackage{xspace}
\usepackage{bm} 
\usepackage{hyperref}
\usepackage{algorithm}
\usepackage[noend]{algpseudocode}
\usepackage{midfloat}
\usepackage{multirow}
\usepackage{booktabs}
\usepackage{siunitx}
\usepackage{float}
\newcommand{\vect}[1]{\mathbf{#1}}

\newcommand{\stt}[1]{{\small\texttt{#1}}}
\newcommand{\davinci}{da~Vinci\textsuperscript\textregistered\xspace}

\usepackage[outdir=./]{epstopdf}

\begin{document}
%
\title{Unsupervised identification of surgical robotic actions from small non-homogeneous datasets}
%
%
%

\author{Daniele Meli,
        Paolo Fiorini
\thanks{Manuscript received: May 16, 2021; Revised: June 30, 2021; Accepted: August 5, 2021.}
\thanks{This paper was recommended for publication by Editor Jessica Burgner-Kahrs upon evaluation of the Associate Editor and Reviewers' comments.}
\thanks{This project has received funding from the European Research Council (ERC) under the European Union's Horizon 2020 research and innovation programme under grant agreement No. 742671 (ARS).}
\thanks{Authors are with Dept. of Computer Science, University of Verona, Italy.
Contact author: Daniele Meli, \texttt{daniele.meli@univr.it}}
\thanks{Digital Object Identifier (DOI): see top of this page.}}

%
%

\markboth{IEEE ROBOTICS AND AUTOMATION LETTERS. PREPRINT VERSION. ACCEPTED AUGUST, 2021}%
{MELI \MakeLowercase{\textit{et al.}}: UNSUPERVISED IDENTIFICATION OF SURGICAL ROBOTIC ACTIONS FROM SMALL NON-HOMOGENEOUS DATASETS}
%



\maketitle

\begin{abstract}
Robot-assisted surgery is an established clinical practice. The automatic identification of surgical actions is needed for a range of applications, including performance assessment of trainees and surgical process modeling for autonomous execution and monitoring. However, supervised action identification is not feasible, due to the burden of manually annotating recordings of potentially complex and long surgical executions. Moreover, often few example executions of a surgical procedure can be recorded. This paper proposes a novel fast algorithm for unsupervised identification of surgical actions in a standard surgical training task, the ring transfer, executed with da Vinci Research Kit. Exploiting kinematic and semantic visual features automatically extracted from a very limited dataset of executions, we are able to significantly outperform state-of-the-art results on a dataset of non-expert executions (58\% vs. 24\% F1-score), and improve performance in the presence of noise, short actions and non-homogeneous workflows, i.e. non repetitive action sequences. 
\end{abstract}


%
\IEEEpeerreviewmaketitle

\section{Introduction}
\label{sec:intro}
%
%
%
%
\IEEEPARstart{R}{obot} assisted minimally invasive surgery (RAMIS) has improved the quality of standard laparoscopic surgery (precise positioning and tremor filtration, visual immersion, faster recovery time for the patient) in a number of clinical scenarios, including urology \cite{palep2009robotic}, esophagectomy \cite{weksler2012robot} and pancreatectomy \cite{daouadi2013robot}. 
Surgical robots as the well established \davinci from Intuitive Surgical are currently employed by novice surgeons for training. 
Automatic quality assessment of trainees by a supervisory system is needed to improve the error-prone verification made by teaching experts. 
A supervisory system will be needed also in the \emph{operating room of the future} \cite{bharathan2013operating}, to improve the safety and efficiency of surgery and assist medical staff in decision making. 
Moreover, a major challenge for the operating room of the future will be the implementation of autonomous robotic surgery to reduce hospital costs and surgeon fatigue. This problem has been only partially addressed, mainly using pre-defined finite state models \cite{nagy2019dvrk,DeRossi2021}, statistical models \cite{Muradore} and logics \cite{ICAR19,IROS2020}.
All the mentioned issues require the definition of an accurate \emph{surgical process model} (SPM) describing the surgical procedure. The definition of the SPM depends on the granularity level of analysis (from low-level description of elementary motory \emph{actions} to high-level sequence of main phases of an intervention) \cite{Lalys}. Though it is possible to define a prior standard SPM from available expert surgical knowledge, learning from surgical practice is needed to build a robust SPM which properly describes variability of the patient's anatomy and the surgical workflow, especially at the action level.
Datasets of RAMIS usually include kinematic records from the robot and videos of the execution.
Given expert annotations of the dataset describing the workflow of execution, Bayes models \cite{charriere2017real} and deep neural networks \cite{twinanda2016endonet, dergachyova2018knowledge} can be successfully used for automatic supervised learning of SPMs.
However, manual annotation of extensive surgical datasets is prone to errors and tedious. 
Current research is mostly focused on the unsupervised identification of actions from surgical datasets. Inspired from advances in human activity recognition \cite{ke2013review,naftel2005classification}, researchers have investigated approaches to unsupervised recognition of surgical actions. \cite{murali2016tsc, shao2018unsupervised} use transition-state clustering (TSC) with deep neural networks on videos and kinematics of JIGSAWS dataset of surgical actions \cite{gao2014jhu}. Approaches based on statistical models, e.g. TSC with Gaussian mixture models \cite{krishnan2018transition}, soft-boundary algorithms with fuzzy clustering scores \cite{fard2016soft} and weakly supervised algorithms \cite{van2019weakly} improve the accuracy. However, they use extensive datasets as JIGSAWS, containing many \emph{homogeneous} repetitions of the same task, i.e. executions which do not differ in the sequence of actions, but present only kinematic variations. 

\begin{figure}[t]
    \centering
    \includegraphics[scale=0.25]{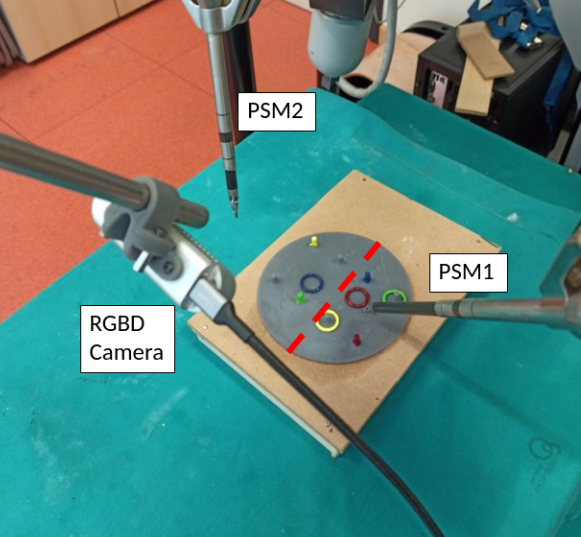}
    \caption{The setup for the ring transfer task. The red dashed line defines reachability regions for the two PSMs.} 
    \label{fig:setup}
\end{figure}

In this paper, we propose a novel algorithm for unsupervised surgical action identification, considering the benchmark training task of ring transfer with da Vinci Research Kit (dVRK) \cite{FLS}. 
The setup for the task is shown in Figure \ref{fig:setup}.
The goal of the task is placing rings on the same-colored pegs, using the two patient-side manipulators (PSMs) of dVRK. Dynamic geometric conditions on the setup affect the sequence of actions. Rings may either be grasped and placed by the same arm or be transferred between arms, for economy of motion. Pegs can be occupied by other rings, so they must be freed before placing another ring. Moreover, rings may be on pegs, thus requiring extraction, or on the base. 

Differently from most state-of-the-art algorithms, 
this paper aims at overcoming some non-addressed limitations of unsupervised surgical action identification, specifically: 1) dealing with small surgical datasets; 2) recognizing actions with short duration; and 3) dealing with non-homogeneous datasets of executions, i.e. differing in the operative conditions and the flow of actions (emulating anatomy-dependent variability in surgery).
To the best of our knowledge, the problem of non-homogeneous datasets has been only partially addressed recently in \cite{shi2021domain}. Authors propose a method based on deep learning from augmented simulated data, and apply it to a peg transfer task similar to ours.
The problem of identifying actions in small datasets of peg transfer has been investigated in \cite{despinoy2015unsupervised}. 

Inspired from the advances in semantic video interpretation from high-level knowledge \cite{akdemir2008ontology}, our algorithm enriches purely kinematic features with semantic domain-dependent features automatically computed from videos of execution. 

The rest of the paper is organized as follows: in Section \ref{sec:materials} we describe our algorithm; in Section \ref{sec:exp} we compare the performance of our algorithm, mainly in comparison with \cite{despinoy2015unsupervised}; finally, Section \ref{sec:conclusion} concludes the paper, in comparison  and outlines possible future research.

\section{Action identification algorithm}
\label{sec:materials}
The input to our action identification algorithm is an \emph{execution trace}, i.e. the kinematic and video stream of an instance of the task; the output is a set of \emph{clustered segments}, i.e. parts of the execution trace corresponding to the same action. 
Ring transfer consists of 6 actions: \stt{move(A,O,C)} which defines motion of one PSM \stt{A} to an object \stt{O} $\in$ \stt{\{ring,peg\}} with color \stt{C}; \stt{move(A,center,C)} which defines the motion of one PSM to the center of the base to transfer a ring to the other arm; \stt{grasp(A,ring,C), release(A)} which define the grasping and releasing action of a colored ring, respectively; \stt{extract(A,ring,C)} which defines the extraction of one ring from a peg where it is placed.

Our algorithm consists of 2 steps: \emph{action segmentation} to identify changepoints delimiting segments in the execution trace; and \emph{action classification} to group together segments corresponding to the same action class.

\subsection{Execution trace}
\label{subsec:trace}
The execution trace of the task consists of a kinematic signature and a synchronized semantic video stream.
\subsubsection{Kinematic signature} It contains relevant kinematic features to describe the trace of execution. Similarly to \cite{despinoy2015unsupervised} chosen as benchmark for this paper, our kinematic signature consists of 16 quantities, including the Cartesian $\vect{p}_A=\{x_{pos,A}, y_{pos,A}, z_{pos,A}\}$ and the quaternion coordinates $\vect{q}_A=\{x_{or,A}, y_{or,A}, z_{or,A}, w_{or,A}\}$ describing the position and orientation of the end effectors of the arms \stt{A} of the dVRK robot; and the opening angles $j_A$ of the grippers. Additional kinematic features may be considered, e.g., speed of the end effectors as in \cite{van2019weakly} and scale-invariant measures of torsion and curvature of the Cartesian trajectories as in \cite{despinoy2015unsupervised}. After preliminary testing with all the features, we decide to omit velocities, curvature and torsion since they do not affect our results. 
\subsubsection{Semantic video stream} It consists of a video stream of the execution trace acquired from a RGB-D camera (calibrated with PSMs as in \cite{roberti2020improving}). Geometric features described in Table \ref{table:SA quantities} are extracted from video frames using standard color segmentation and shape recognition (to distinguish between pegs and rings) with Random Sample Consensus \cite{ransac}. Then, they are combined with kinematic features and translated to semantic features called \emph{fluents} with the situation awareness (SA) algorithm proposed in \cite{IROS2020}, according to the following relations:
\begin{align*}
    \stt{at(A,ring,C)} &\leftarrow ||\vect{p}_A - \vect{p}_{rc}||_2 < rr\\
    \stt{at(A,peg,C)} &\leftarrow ||\vect{p}_A - \vect{p}_{pc}||_2 < rr \ \land \\ &\land \ z_{pc} < z_{pos,A}\\
    \stt{in\_hand(A,ring,C)} &\leftarrow ||\vect{p}_A - \vect{p}_{rc}||_2 < rr \ \land \ j_A < \frac{\pi}{8} \footnotemark\\
    \stt{on(ring,C1,peg,C2)} &\leftarrow ||\vect{p}_{rc1} - \vect{p}_{pc2}||_2 < rr \ \land \\ &\land \ z_{rc1} < z_{pc2}\\
    \stt{reachable(A$_x$,O,C)} &\leftarrow argmin_{A}|y_{oc} - y_{pos,A}| = A_x\\
    \stt{closed\_gripper(A)} &\leftarrow j_A < \frac{\pi}{8}\\
    \stt{at(A,center)} &\leftarrow ||\{x_{pos,A}, y_{pos,A}\} - \{x_{b}, y_{b}\}||_2 <\\ &< rr
\end{align*}
\footnotetext{$\frac{\pi}{8}$ is chosen empirically to identify that the gripper is closed, since the angle is not $0$ when the PSM is holding a ring.}
\begin{table}[t]
\caption{Geometric features detected in the frames of the video stream of the execution traces by the SA algorithm.}
\label{table:SA quantities}
\centering
\begin{tabular}{ll}
\toprule
\textbf{Name} & \textbf{Description}\\
\midrule
$\vect{p}_{rc} = \{x_{rc}, y_{rc}, z_{rc}\}$ & position of ring with color \stt{C} (center)\\
$\vect{p}_{pc} = \{x_{pc}, y_{pc}, z_{pc}\}$ & position of peg with color \stt{C} (tip)\\
$\vect{p}_{b} = \{x_{b}, y_{b}, z_{b}\}$ & position of center of peg base \\
$rr$ & ring radius\\
\bottomrule
\end{tabular}
\end{table}
The semantics of fluents is clear. In particular, \stt{reachable(A,O,C)} defines which objects are in the reachability region of each PSM, as delimited by the red dashed line in Figure \ref{fig:setup}.

\subsection{Action segmentation}
Given an execution trace, the segmentation algorithm \ref{alg:seg} identifies changepoints corresponding to starting / ending timesteps of actions of the task. 
The algorithm involves two main steps: changepoint detection from kinematics, and changepoint filtering from the semantic video stream.
\subsubsection{Changepoint detection} The features in the kinematic signature are first normalized by their maximum values to allow comparison between different configurations of the setup for the task (different locations of objects in the scene and distances to the PSMs). Then, filtering at \SI{1.5}{\hertz} is applied to each feature, to consider only relevant motions within the fundamental frequency of human gestures \cite{loram2006frequency}. We identify changepoints in the kinematic signature evaluating peaks in the 2nd derivative of the features, and only peaks above a percentage $\alpha$ of the maximum absolute value are considered. In order to remove implicit noise in the 2nd derivative, we compute it using Savitzky-Golay filter (SGF) \cite{5888646}, a digital filter based on polynomial fitting over an adjustable time window. SGF captures higher-order momenta in the data, while removing noise in the signal. SGF has a long history of successful applications in time series analysis, e.g. for kinematic analysis of the human arm \cite{wang1999three} and neural signals \cite{pistohl2008prediction}. Notice that there exist several algorithms and implementations for changepoint detection, and an extensive review can be found in \cite{truong2020selective}. We choose peak detection from SGF on the 2nd derivative of kinematic features because it is fast and effective for our experiments, and it is robust since it requires no prior assumption on the number of segments, and few parameters to be tuned (time window set to 21 timesteps and $\alpha=20\%$ empirically).

\subsubsection{Changepoint filtering} Changepoint detection may return spurious changepoints, corresponding e.g. to abrupt motions during the execution. Then, we reduce the set of changepoints from the detection algorithm, excluding consecutive changepoints distant less than $1 s$ (average duration of shortest actions, see Section \ref{subsec:testA}). Then, we compute fluents from the video stream in correspondence of each changepoint, and we omit consecutive changepoints which share the same set of fluents. This choice is made under the assumption that \emph{different actions derive from different environmental conditions}: hence, two consecutive changepoints defining the start of two different actions shall not share the same set of fluents. 
Our approach requires tuning of fewer parameters than the one proposed in \cite{despinoy2015unsupervised}. In fact, authors of \cite{despinoy2015unsupervised} propose to identify changepoints using persistence analysis \cite{edelsbrunner2000topological}. Persistence analysis removes noise and identifies relevant local minima and maxima in the kinematic features, but requires the definition of an ad-hoc threshold for each feature, which is often task- and operator-dependent.
\begin{algorithm}[t]
    \caption{Action segmentation algorithm}\label{alg:seg}
    \begin{algorithmic}[1]
        \State \textbf{Input}: Execution trace with temporal kinematic signature $K(t)$ and video stream $V(t)$, threshold for peak detection $\alpha$
        \State \textbf{Output}: Set of changepoints $C$
        \State \textbf{Initialize}: Normalize and filter $K(t)$, $C = []$
        \State \% \textit{Changepoint detection}
        \For{$k(t)$ kinematic feature $\in K(t)$}
            \State $\ddot{k}(t)$ = SGF($k(t)$)
            \State peaks = PeakDetect($\ddot{k}(t)$)
            \For{$p \in$ peaks}
                \If{$|\ddot{k}(p)| > \alpha \cdot \max|\ddot{k}(p)|$}
                    \State $C$.append($p$)
                \EndIf
            \EndFor
        \EndFor
        \State \% \textit{Changepoint filtering}
        \State fluents $F = []$
        \State $c_{old} = 0$
        \For{$c \in C$}
            \State old\_fluents $F_{old}= F$
            \State $F$ = FluentCompute($V(c)$)
            \If{$F == F_{old} \ \lor \ c - c_{old} < 1s$}
                \State $C$.remove($c$)
            \EndIf
            \State $c_{old} = c$
        \EndFor
        \State \Return{$C$}
    \end{algorithmic}
\end{algorithm}

\subsection{Action classification}
\label{subsec:cluster}
Given the changepoints of the execution trace(s), we use $k$-NN classification to group segments corresponding to same actions. $k$-NN is preferred to other classification methods as support vector machines \cite{despinoy2015unsupervised} and self-organizing maps \cite{naftel2005classification} because it usually performs better on small datasets.
We associate each segment in the execution trace with a feature vector $\vect{f} = [\vect{f}_1, \vect{f}_2, \vect{f}_3]$, containing both kinematic information as in \cite{despinoy2015unsupervised} and semantic environmental information from the video stream.
\subsubsection{$\vect{f}_1$}
It is an array of reals representing the kinematic features of each segment. Each feature $k_i$ in the kinematic signature of the segment is shifted to start from $t=0$ (for fair comparison between different segments of the same action); then, it is approximated by a polynomial $p_i(t) = \sum_{j=1}^n a_{j,i} t^j$, with $n \in \mathbb{N}$ arbitrary degree as in \cite{despinoy2015unsupervised} ($n=5$ empirically in this work). Then, $\vect{f}_1$ is built concatenating coefficients $a_{j,i}$, resulting in an array of dimension $(n+1) \cdot 16$. 
Polynomial approximation is not the only option for kinematic representation. For instance, in \cite{naftel2005classification} Fourier coefficients are used for noise-robust action identification in the CAVIAR dataset for human gestures \cite{crowley2004context}. However, in our experiments Fourier approximation does not show to improve the performance, so it is discarded.
We want our classification algorithm to generalize over the actual PSM executing a specific action: for instance, two segments corresponding to \stt{move(PSM1,ring,red), move(PSM2,ring,yellow)} must be classified in the same cluster, corresponding to the abstract action \stt{move(A,ring,C)}. Hence, we check whether one of the PSMs does not move, i.e. the kinematic signatures at the starting and ending changepoints have no significant difference. In case only one arm moves, the coefficients of the polynomial approximations of the kinematic features for that specific arm are added to $\vect{f}_1$, and null coefficients for the other arm are appended to complete the vector. In case both PSMs are moving, coefficients of PSM1 and PSM2 are concatenated to build $\vect{f}_1$. In this way, there is no distinction whether an action is executed by one arm or the other.
\subsubsection{$\vect{f}_2$}
It is a Boolean array representing fluents holding at the beginning changepoint of each segment. Each entry in the array corresponds to a fluent defined in Section \ref{subsec:trace}, excluding \stt{reachable} which is identified at all timesteps in the video stream (all rings and pegs are always reachable at least by one arm), hence it is neglected. Each entry in the array has true value if the corresponding fluent holds at the beginning of the segment, otherwise it is set to false. In order to guarantee the invariance of the classification algorithm with respect to the arm and color instances, $\vect{f}_2$ has two entries for each fluent, one per PSM, and the color attribute is ignored. Similarly to the generation of $\vect{f}_1$, if a fluent holds only for one arm, then the first corresponding entry is set to true, regardless of the specific arm. The final dimension of $\vect{f}_2$ is 12.
\subsubsection{$\vect{f}_3$}
It is a 16D Boolean vector with entries corresponding to each feature in the kinematic signature of the segment. If one signature varies from the beginning to the ending changepoint, the value of the corresponding entry is set to true (as for $\vect{f}_1$, the order of entries for the two arms does not depend on the specific PSM). $\vect{f}_3$ is needed because instances of the same action may significantly differ from a kinematic perspective, due to the different relative position of objects and arms in the scene (for instance, in the scenario depicted in Figure \ref{fig:setup}, \stt{move(PSM1,ring,red)} requires a completely different motion than the one needed for \stt{move(PSM2,ring,blue)}). However, we expect that the same kinematic features vary along the segment.
\subsubsection{Classification algorithm}
$k$-NN classification compares feature vectors for different segments in the execution trace(s) containing both Boolean and real values. Hence, we need to define a mixed distance metric $d_{i,l}$ between segments $i,l$:
\begin{equation}
    \label{eq:metric}
    d_{i,l} = \sqrt{\frac{d_e^2}{d_{emax}^2} + \frac{d_h^2}{d_{hmax}^2}}
\end{equation}
where $d_e$ is the standard Euclidean distance between coefficients in $\vect{f}_1$ for the two segments, $d_e = \sqrt{\sum_{j=1}^n (a_{j,i} - a_{j,l})^2}$; while $d_h$ is the Hamming distance between $[\vect{f}_2, \vect{f}_3]$ for the two segments
\begin{equation*}
    d_h = \frac{h_{i,l}}{\dim([\vect{f}_2, \vect{f}_3]_i)}
\end{equation*}
being $h_{i,l}$ the number of different values in $[\vect{f}_2, \vect{f}_3]$ for segment $i$ with respect to segment $l$, and $\dim(\cdot)$ the dimension of an array. $d_{emax}, d_{hmax}$ are normalizing factors which are needed for fair comparison of Euclidean and Hamming distances, and they are chosen as the maximum cost resulting from $k$-NN classification with only Euclidean distance over $\vect{f}_1$ and only Hamming distance over $[\vect{f}_2, \vect{f}_3]$, respectively.
$k$ for $k$-NN classification is chosen as the number of occurrencies of the most frequent action in the dataset of executions. In this way, the algorithm is able to recognize all instances of actions in the dataset. A higher $k$ value would also be valid; however, in the experiments we show that the best classification performance is achieved with the minimum $k$.

\section{Experiments}
\label{sec:exp}
We prove the advantages of our unsupervised action identification algorithm with three different experiments. In Test A, we consider nine executions of the ring transfer task in homogeneous \emph{standard environmental condition}, i.e. with all rings placed on grey pegs and requiring transfer between arms. 
The task is executed by 3 users (not surgeons) with different expertise in using the dVRK.
This experiment is useful for comparison with benchmark \cite{despinoy2015unsupervised}, which considers multiple homogeneous executions of the same task by three expert users.

In Test B, we first consider an execution in standard environmental conditions, performed autonomously as explained in \cite{IROS2020}, where the robot's trajectories were described as Dynamic Movement Primitives (DMPs) \cite{Sch06,GMCDSF19,Ginesi2021,ginesi2019dmp}. 
DMPs allow to learn and replicate the shape of trajectories executed by humans, both in Cartesian and orientation space. In \cite{IROS2020}, 15 executions of the ring transfer task were recorded from the same users of Test A and used for learning trajectories of each action in the task. Hence, though the execution is autonomous, it reproduces the kinematics of human executions.
In this test, we generate synthetic task replications from the original autonomous execution, adding low-frequency human-like kinematic noise to test the robustness of our algorithm with respect to noise (hence, the dataset is still homogeneous). 

\begin{figure*}[t]
    \centering
    \begin{subfigure}{0.3\textwidth}
    \includegraphics[height = 0.55\linewidth]{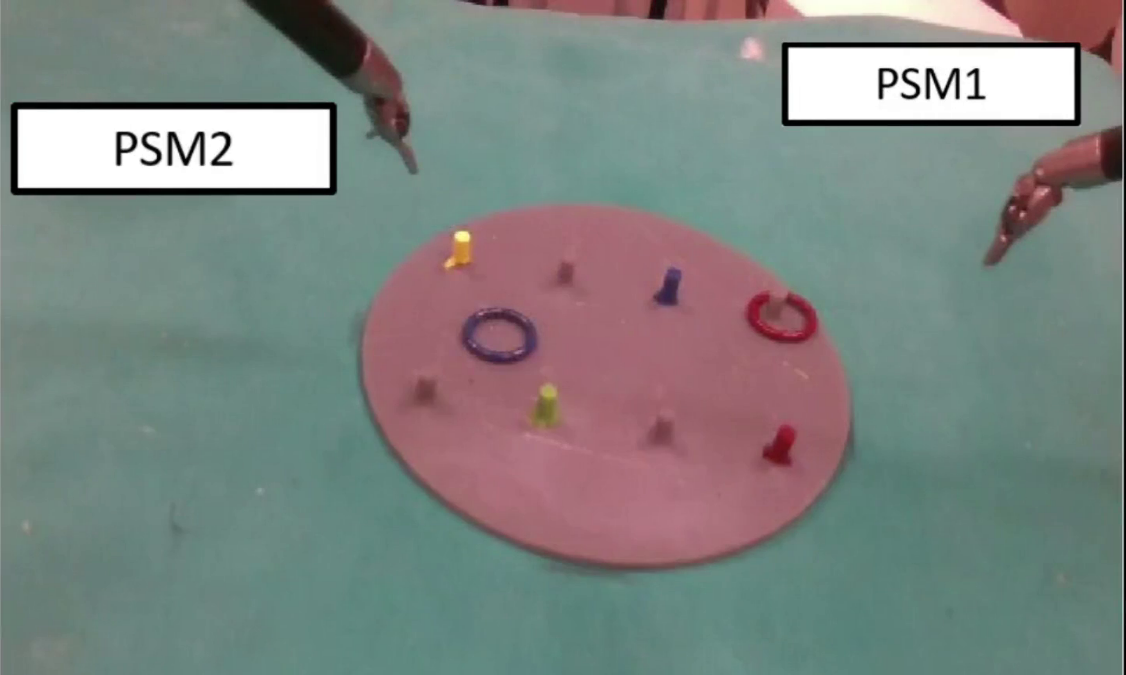}
    \caption{Execution with failure (fallen ring)}
    \label{subfig:exec1}
    \end{subfigure}
    \begin{subfigure}{0.3\textwidth}
    \includegraphics[height = 0.55\linewidth]{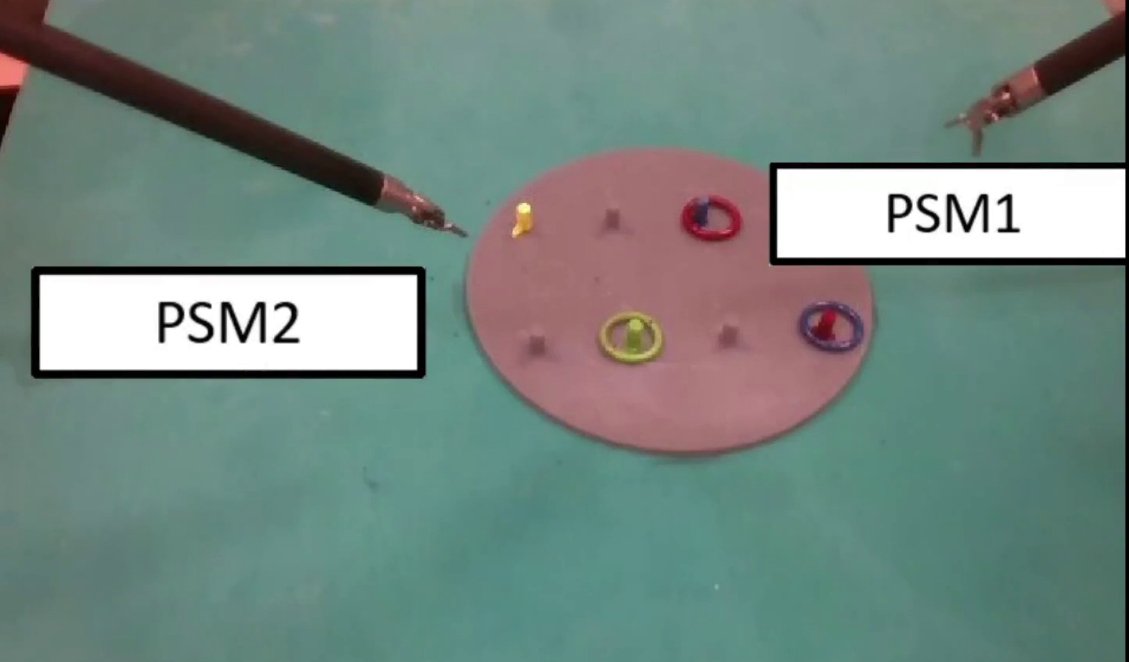}
    \caption{Execution with occupied pegs}
    \label{subfig:exec2}
    \end{subfigure}
    \begin{subfigure}{0.3\textwidth}
    \includegraphics[height = 0.55\linewidth]{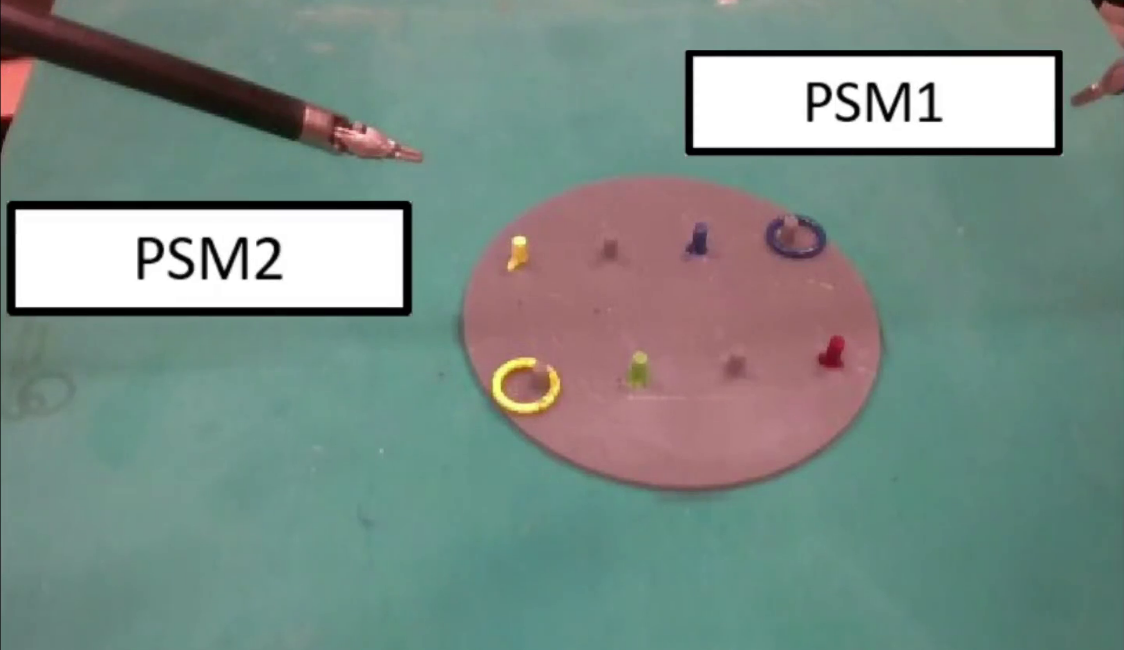}
    \caption{Simultaneous execution by two PSMs}
    \label{subfig:exec3}
    \end{subfigure}
    \caption{Initial environmental conditions for the task executions used in Test B.}
    \label{fig:execs}
\end{figure*}

Finally, in Test C, we consider a non-homogeneous dataset consisting of only the standard execution of Test B without noise, and three autonomous executions of the task in unconventional environmental conditions, shown in Figure \ref{fig:execs}.
All execution traces (kinematic readings from dVRK\footnote{\url{https://github.com/jhu-dvrk/sawIntuitiveResearchKit}} and point clouds from a Realsense D435 RGB-D camera\footnote{\url{https://www.intelrealsense.com/depth-camera-d435/}}) are collected with the Robot Operating System to ensure synchronization.

\begin{figure*}[t]
    \centering
    \includegraphics[width = 0.8\linewidth]{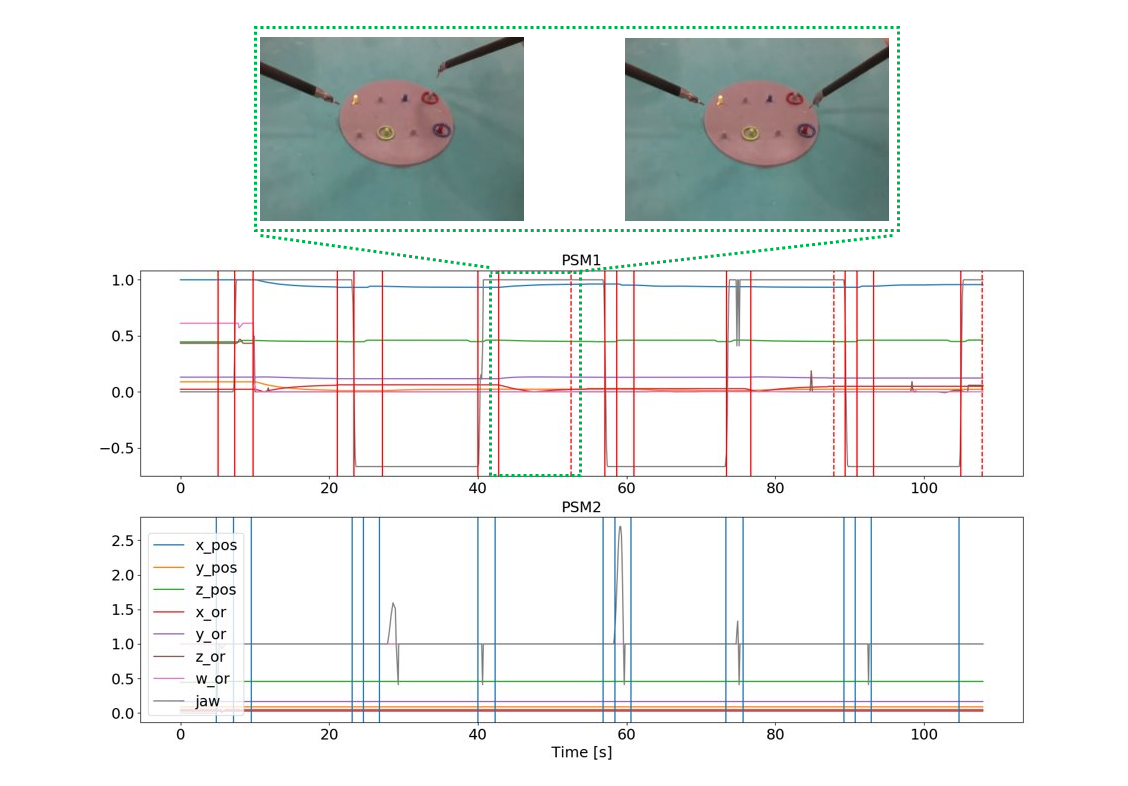}
    \caption{Kinematic signature (after filtering) and segmentation results for the execution in Figure \ref{subfig:exec2}. Blue vertical lines represent real changepoints, red solid lines represent changepoints identified by our algorithm, red dashed lines represent changepoints excluded after fluent detection.
    Frames on the top correspond to two consecutive changepoints sharing the same set of fluents (\stt{on(ring,red,peg,grey), on(ring,blue,peg,red)} and the reachability fluents), so the second one is omitted.}
    \label{fig:blue_red}
\end{figure*}
For a qualitative understanding especially of our changepoint filtering algorithm, Figure \ref{fig:blue_red} shows the results of our segmentation algorithm applied to the execution depicted in Figure \ref{subfig:exec2}. Our changepoint detection algorithm finds more changepoints (red solid lines) than real ones from manual segmentation (blue lines). However, we are able to filter out wrong changepoints (red dashed lines) exploiting fluent detection from video stream, when two consecutive changepoints share the same set of fluents. 

We quantify the performance of our algorithm using the reference metrics proposed in \cite{despinoy2015unsupervised}. Specifically, we rate the segmentation accuracy using the \emph{matching score}:
\begin{equation*}
    M_i = \frac{|\cap(t_i, g_i)|}{|g_i|}
\end{equation*}
where $t_i$ is the segment identified by our algorithm, $g_i$ is the real segment corresponding to the $i$-th action, and $|\cdot|$ denotes the temporal length of a segment. The matching score measures the overlapping between the identified and actual segment, normalized by the length of the actual segment. 
The results of action classification are quantified using \emph{precision, recall} and \emph{F1-score}. Precision for an action class $A$ is defined as:
\begin{equation*}
    Pr = \frac{TP}{FP+TP}
\end{equation*}
being $TP$ the number of true positives, i.e. segments correctly classified in $A$, and $FP$ the number of false positives, i.e. segments mistakenly classified in $A$. Precision measures the rate of success of the classification algorithm on the full dataset of executions. 
Since $k$ for $k$-NN classification is chosen as the maximum number of occurrencies of the most frequent action in the dataset, $FP+TP$ is chosen for the $i$-th action class as $\dim(P)$, where $P$ is the maximum set of segments such that:  
\begin{itemize}
    \item $\dim(P) \geq n_{occ,i}$, where $n_{occ,i}$ is the number of occurrencies of the $i$-th action in the dataset;
    \item the score of the last segment in $P$ is $s_P = s_{max}$, being $s_{max}$ the score of the $n_{occ,i}$-th segment in $P$.
\end{itemize}
In this way, the scores for less frequent actions are not significantly affected by the most frequent ones.
Recall for an action class $A$ is defined as:
\begin{equation*}
    Rec = \frac{TP}{FN+TP}
\end{equation*}
with $FN$ the number of false negatives, i.e. segments mistakenly not included in $A$. Recall measures the rate of success of the classification algorithm for a single action. F1-score combines precision and recall as follows:
\begin{equation*}
    F1 = 2\frac{Pr \cdot Rec}{Pr + Rec}
\end{equation*}

\subsection{Test A: Homogeneous dataset from human executions}
\label{subsec:testA}
\begin{figure}[t]
    \centering
    \includegraphics[scale=0.19]{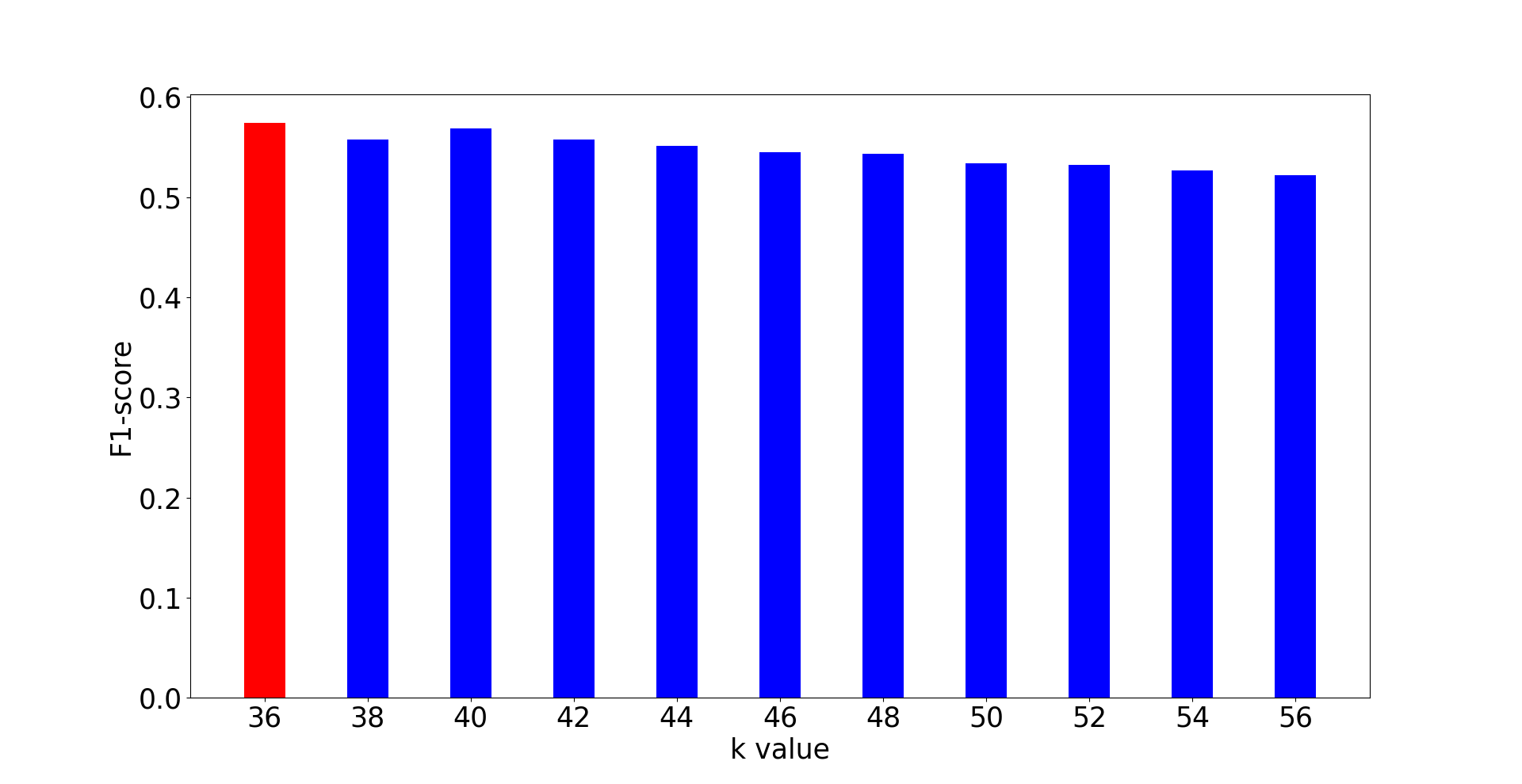}
    \caption{Average F1-score over all actions for Test A, with increasing $k$ value. The minimum value (red bar) is the optimal one.} 
    \label{fig:k_analysis}
\end{figure}

In this test, we consider 9 executions of the ring transfer task in standard environmental conditions, executed by 3 users with different expertise in using the dVRK. 
Gripper actions and \stt{extract} last $1.05 \pm 1.09$ \si{\s}, while \stt{move} actions last $3.69 \pm 2.90$ \si{\s} (comparably with peg transfer in \cite{despinoy2015unsupervised} on average). The high variance is due to the different expertise of users.
Each execution trace contains 36 actions (actions appear multiple times in each execution).
Table \ref{tab:matchA} reports the matching scores (for each action and on average over all actions). The segmentation results are slightly better than \cite{despinoy2015unsupervised}, though comparable.
\begin{table}[t]
\caption{Matching scores for Test A. All values are percentages.}
\label{tab:matchA}
\begin{tabular*}{\hsize}{@{\extracolsep{\fill}}lll@{}}
\toprule
\textbf{Action} & & \textbf{Matching score}\\
\midrule
\stt{move(A,ring,C)} & & 85.65\\
\stt{move(A,peg,C)} & & 90.69\\
\stt{move(A,center,C)} & & 82.43\\
\stt{grasp(A,ring,C)} & & 77.05\\
\stt{extract(A,ring,C)} & & 82.73\\
\stt{release(A)} & & 90.18\\
\midrule
\textbf{Average} & \begin{tabular}{@{}l@{}} Our method \\ Method in \cite{despinoy2015unsupervised} \end{tabular} & \begin{tabular}{@{}l@{}} \textbf{84.79} \\ 81.90 \end{tabular}\\
\bottomrule
\end{tabular*}
\end{table}
In Table \ref{tab:clusterA}, we report the results of our classification algorithm. 
We choose $k=36$ for $k$-NN classification, i.e. the number of occurrencies of the most frequent action (\stt{release}) in the dataset. 
As explained in Section \ref{subsec:cluster}, this is the minimum possible value for $k$ to recognize at least all action instances.
However, there is no prior guarantee that this is the optimal value for the classification performance.
Hence, in Figure \ref{fig:k_analysis} we run a preliminary test wih increasing values of $k \in [36, 56]$ with a step of 2, and we compare the average F1-score over all actions in the dataset to identify the best choice of $k$. Results show that $k=36$ is the optimal value. Hence, also in the following experiments we will choose $k$ as the minimum possible value.
Given $k$, we first use the full feature array $[\vect{f}_1, \vect{f}_2, \vect{f}_3]$ presented in Section \ref{subsec:cluster} as input to the classification algorithm. However, this results in poor performance, especially for short actions. Hence, we use only Boolean features $[\vect{f}_2, \vect{f}_3]$ for short actions to obtain the results in Table \ref{tab:clusterA}. Results are significantly improved with respect to \cite{despinoy2015unsupervised}, whose method suffers from the executions by non-expert users, on the contrary. 
\begin{table}[t]
\caption{Action classification results for Test A. All values are percentages.}
\label{tab:clusterA}
\begin{tabular*}{\hsize}{@{\extracolsep{\fill}}lllll@{}}
\toprule
\textbf{Action} & \textbf{Feature array} & $Pr$ & $Rec$ & $F1$\\
\midrule
\stt{move(A,ring,C)} & \begin{tabular}{@{}l@{}} $[\vect{f_1},\vect{f_2},\vect{f_3}]$ \\ $\vect{f_1}$ as in \cite{despinoy2015unsupervised} \end{tabular} & \begin{tabular}{@{}l@{}} \textbf{100.00} \\ 22.22 \end{tabular} & \begin{tabular}{@{}l@{}} \textbf{83.33} \\ 22.22 \end{tabular} & \begin{tabular}{@{}l@{}} \textbf{90.91} \\ 22.22 \end{tabular}\\
\midrule
\stt{move(A,peg,C)} & \begin{tabular}{@{}l@{}} $[\vect{f_1},\vect{f_2},\vect{f_3}]$ \\ $\vect{f_1}$ as in \cite{despinoy2015unsupervised} \end{tabular} & \begin{tabular}{@{}l@{}} \textbf{55.56} \\ 19.44 \end{tabular} & \begin{tabular}{@{}l@{}} \textbf{55.56} \\ 19.44 \end{tabular} & \begin{tabular}{@{}l@{}} \textbf{55.56} \\ 19.44 \end{tabular}\\
\midrule
\stt{move(A,center,C)} & \begin{tabular}{@{}l@{}} $[\vect{f_1},\vect{f_2},\vect{f_3}]$ \\ $\vect{f_1}$ as in \cite{despinoy2015unsupervised} \end{tabular} & \begin{tabular}{@{}l@{}} \textbf{58.33} \\ 25.00 \end{tabular} & \begin{tabular}{@{}l@{}} \textbf{58.33} \\ 25.00 \end{tabular} & \begin{tabular}{@{}l@{}} \textbf{58.33} \\ 25.00 \end{tabular}\\
\midrule
\stt{grasp(A,ring,C)} & \begin{tabular}{@{}l@{}} $[\vect{f_2},\vect{f_3}]$ \\ $\vect{f_1}$ as in \cite{despinoy2015unsupervised} \end{tabular} &  \begin{tabular}{@{}l@{}} \textbf{41.23} \\ 17.78 \end{tabular} & \begin{tabular}{@{}l@{}} \textbf{40.00} \\ 22.22 \end{tabular} & \begin{tabular}{@{}l@{}} \textbf{40.61} \\ 19.66 \end{tabular}\\
\midrule
\stt{extract(A,ring,C)} & \begin{tabular}{@{}l@{}} $[\vect{f_2},\vect{f_3}]$ \\ $\vect{f_1}$ as in \cite{despinoy2015unsupervised} \end{tabular} & \begin{tabular}{@{}l@{}} \textbf{66.67} \\ 30.56 \end{tabular} & \begin{tabular}{@{}l@{}} \textbf{66.67} \\ 30.56 \end{tabular} & \begin{tabular}{@{}l@{}} \textbf{66.67} \\ 30.56 \end{tabular}\\
\midrule
\stt{release(A)} & \begin{tabular}{@{}l@{}} $[\vect{f_2},\vect{f_3}]$ \\ $\vect{f_1}$ as in \cite{despinoy2015unsupervised} \end{tabular} & \begin{tabular}{@{}l@{}} \textbf{40.00} \\ 26.67 \end{tabular} & \begin{tabular}{@{}l@{}} \textbf{40.00} \\ 26.67 \end{tabular} & \begin{tabular}{@{}l@{}} \textbf{40.00} \\ 26.67 \end{tabular}\\
\midrule
\textbf{Average} & \begin{tabular}{@{}l@{}} Our method \\ $\vect{f_1}$ as in \cite{despinoy2015unsupervised} \end{tabular} & \begin{tabular}{@{}l@{}} \textbf{60.30} \\ 23.61 \end{tabular} & \begin{tabular}{@{}l@{}} \textbf{57.32} \\ 24.35 \end{tabular} & \begin{tabular}{@{}l@{}} \textbf{58.68} \\ 23.93 \end{tabular}\\
\bottomrule
\end{tabular*}
\end{table}

\subsection{Test B: Homogeneous dataset with noise}
In this test, we consider a dataset consisting of a task execution in standard conditions learned from humans with DMPs, and nine more executions generated adding low-frequency noise to the kinematic signature. This generates an homogeneous dataset with noise, which increases the kinematic variability between different users and requires additional robustness. Noise is generated with power frequency spectrum:
\begin{equation*}
    S(f) = \beta \frac{1}{f^\lambda}
\end{equation*}
with $\lambda = 7.5$ so that frequencies above the fundamental frequency \SI{1.5}{\hertz} of human hand \cite{loram2006frequency} have power below 0.05$\beta$, hence they can be neglected. $\beta$ is related to the noise variance. We vary it in the range $[0.01, 0.09]$ with a step of 0.01 to generate the synthetic executions. 
In Table \ref{tab:matchB} we evaluate the matching score considering the full synthetic dataset. The average matching score over all actions with our Algorithm \ref{alg:seg} improves the results of \cite{despinoy2015unsupervised}.
\begin{table}[t]
\caption{Matching scores for Test B. All values are percentages.}
\label{tab:matchB}
\begin{tabular*}{\hsize}{@{\extracolsep{\fill}}lll@{}}
\toprule
\textbf{Action} & & \textbf{Matching score}\\
\midrule
\stt{move(A,ring,C)} & & 94.78\\
\stt{move(A,peg,C)} & & 90.47\\
\stt{move(A,center,C)} & & 97.53\\
\stt{grasp(A,ring,C)} & & 76.02\\
\stt{extract(A,ring,C)} & & 71.80\\
\stt{release(A)} & & 93.18\\
\midrule
\textbf{Average} & \begin{tabular}{@{}l@{}} Our method \\ Method in \cite{despinoy2015unsupervised} \end{tabular} & \begin{tabular}{@{}l@{}} \textbf{87.30} \\ 81.90 \end{tabular}\\
\bottomrule
\end{tabular*}
\end{table}
We then evaluate the performance of our action classification algorithm in Table \ref{tab:clusterB}.
We choose $k=36$ (number of occurrencies of \stt{release} action) for $k$-NN classification. Results are comparable with Test A, though the improvement with respect to \cite{despinoy2015unsupervised} is significant only for short actions.
\begin{table}[t]
\caption{Action classification results for Test B. All values are percentages.}
\label{tab:clusterB}
\begin{tabular*}{\hsize}{@{\extracolsep{\fill}}lllll@{}}
\toprule
\textbf{Action} & \textbf{Feature array} & $Pr$ & $Rec$ & $F1$\\
\midrule
\stt{move(A,ring,C)} & \begin{tabular}{@{}l@{}} $[\vect{f_1},\vect{f_2},\vect{f_3}]$ \\ $\vect{f_1}$ as in \cite{despinoy2015unsupervised} \end{tabular} & \begin{tabular}{@{}l@{}} 75.00 \\ 75.00 \end{tabular} & \begin{tabular}{@{}l@{}} 75.00 \\ 75.00 \end{tabular} & \begin{tabular}{@{}l@{}} 75.00 \\ 75.00 \end{tabular}\\
\midrule
\stt{move(A,peg,C)} & \begin{tabular}{@{}l@{}} $[\vect{f_1},\vect{f_2},\vect{f_3}]$ \\ $\vect{f_1}$ as in \cite{despinoy2015unsupervised} \end{tabular} & \begin{tabular}{@{}l@{}} 100.00 \\ 100.00 \end{tabular} & \begin{tabular}{@{}l@{}} 100.00 \\ 100.00 \end{tabular} & \begin{tabular}{@{}l@{}} 100.00 \\ 100.00 \end{tabular}\\
\midrule
\stt{move(A,center,C)} & \begin{tabular}{@{}l@{}} $[\vect{f_1},\vect{f_2},\vect{f_3}]$ \\ $\vect{f_1}$ as in \cite{despinoy2015unsupervised} \end{tabular} & \begin{tabular}{@{}l@{}} 40.82 \\ \textbf{45.00} \end{tabular} & \begin{tabular}{@{}l@{}} \textbf{55.56} \\ 50.00 \end{tabular} & \begin{tabular}{@{}l@{}} 47.06 \\ \textbf{47.37} \end{tabular}\\
\midrule
\stt{grasp(A,ring,C)} & \begin{tabular}{@{}l@{}} $[\vect{f_2},\vect{f_3}]$ \\ $\vect{f_1}$ as in \cite{despinoy2015unsupervised} \end{tabular} &  \begin{tabular}{@{}l@{}} \textbf{40.00} \\ 17.78 \end{tabular} & \begin{tabular}{@{}l@{}} \textbf{50.00} \\ 22.22 \end{tabular} & \begin{tabular}{@{}l@{}} \textbf{44.44} \\ 19.66 \end{tabular}\\
\midrule
\stt{extract(A,ring,C)} & \begin{tabular}{@{}l@{}} $[\vect{f_2},\vect{f_3}]$ \\ $\vect{f_1}$ as in \cite{despinoy2015unsupervised} \end{tabular} & \begin{tabular}{@{}l@{}} \textbf{51.02} \\ 47.37 \end{tabular} & \begin{tabular}{@{}l@{}} \textbf{69.44} \\ 50.00 \end{tabular} & \begin{tabular}{@{}l@{}} \textbf{58.82} \\ 48.65 \end{tabular}\\
\midrule
\stt{release(A)} & \begin{tabular}{@{}l@{}} $[\vect{f_2},\vect{f_3}]$ \\ $\vect{f_1}$ as in \cite{despinoy2015unsupervised} \end{tabular} & \begin{tabular}{@{}l@{}} \textbf{40.00} \\ 26.67 \end{tabular} & \begin{tabular}{@{}l@{}} \textbf{40.00} \\ 26.67 \end{tabular} & \begin{tabular}{@{}l@{}} \textbf{40.00} \\ 26.67 \end{tabular}\\
\midrule
\textbf{Average} & \begin{tabular}{@{}l@{}} Our method \\ $\vect{f_1}$ as in \cite{despinoy2015unsupervised} \end{tabular} & \begin{tabular}{@{}l@{}} \textbf{57.81} \\ 51.97 \end{tabular} & \begin{tabular}{@{}l@{}} \textbf{65.00} \\ 53.98 \end{tabular} & \begin{tabular}{@{}l@{}} \textbf{60.89} \\ 52.89 \end{tabular}\\
\bottomrule
\end{tabular*}
\end{table}

\subsection{Test C: Non-homogeneous dataset}
In Test C we consider a dataset of only 4 executions under non-homogeneous environmental conditions, i.e. the initial conditions of the setup and the action sequence vary between executions. One execution is in standard environmental conditions (36 actions), while 3 other ones are depicted in Figure \ref{fig:execs} and include a failure condition with 18 actions (the ring falls during the execution, Figure \ref{subfig:exec1}), an unconventional scenario with occupied colored pegs involving 17 actions \ref{subfig:exec2}, and an execution with PSMs moving simultaneously \ref{subfig:exec3} including 12 actions. Actions in each execution repeat multiple times. This is a challenging dataset for unsupervised action identification, since not all actions appear in all executions, and both kinematic and semantic features do not repeat in the same way over the whole dataset. 
Table \ref{tab:matchC} shows that the matching score is comparable with Test B both for each action and on average, and higher than \cite{despinoy2015unsupervised} on average.
Results of $k$-NN classification with $k=21$ (number of occurrencies of \stt{release} action) are shown in Table \ref{tab:clusterC}. Our method outperforms the scores in \cite{despinoy2015unsupervised} (average F1-score rises from 54\% to almost 67\%). The results for \stt{extract} action are significantly improved, reaching precision of 100\% and F1-score of 77\% (both scores reach only 12\% on the dataset of \cite{despinoy2015unsupervised}). Also F1-score and precision (almost double) for \stt{move(A,center,C)} are better. This is even more significant considering that this action is the least frequent in the dataset, appearing only in two scenarios (once in Figure \ref{subfig:exec1} and 4 times under standard environmental conditions). A slight decrease in the performance is only recorded for \stt{move(A,ring,C)}, though the F1-score is the highest among all actions.
\begin{table}[t]
\caption{Matching scores for Test C. All values are percentages.}
\label{tab:matchC}
\begin{tabular*}{\hsize}{@{\extracolsep{\fill}}lll@{}}
\toprule
\textbf{Action} & & \textbf{Matching score}\\
\midrule
\stt{move(A,ring,C)} & & 95.36\\
\stt{move(A,peg,C)} & & 92.83\\
\stt{move(A,center,C)} & & 96.35\\
\stt{grasp(A,ring,C)} & & 78.40\\
\stt{extract(A,ring,C)} & & 83.08\\
\stt{release(A)} & & 85.45\\
\midrule
\textbf{Average} & \begin{tabular}{@{}l@{}} Our method \\ Method in \cite{despinoy2015unsupervised} \end{tabular} & \begin{tabular}{@{}l@{}} \textbf{88.58} \\ 81.90 \end{tabular}\\
\bottomrule
\end{tabular*}
\end{table}
\begin{table}[t]
\caption{Action classification results for Test C. All values are percentages.}
\label{tab:clusterC}
\begin{tabular*}{\hsize}{@{\extracolsep{\fill}}lllll@{}}
\toprule
\textbf{Action} & \textbf{Feature array} & $Pr$ & $Rec$ & $F1$\\
\midrule
\stt{move(A,ring,C)} & \begin{tabular}{@{}l@{}} $[\vect{f_1},\vect{f_2},\vect{f_3}]$ \\ $\vect{f_1}$ as in \cite{despinoy2015unsupervised} \end{tabular} & \begin{tabular}{@{}l@{}} 81.82 \\ \textbf{90.00} \end{tabular} & \begin{tabular}{@{}l@{}} 90.00 \\ 90.00 \end{tabular} & \begin{tabular}{@{}l@{}} 85.72 \\ \textbf{90.00} \end{tabular}\\
\midrule
\stt{move(A,peg,C)} & \begin{tabular}{@{}l@{}} $[\vect{f_1},\vect{f_2},\vect{f_3}]$ \\ $\vect{f_1}$ as in \cite{despinoy2015unsupervised} \end{tabular} & \begin{tabular}{@{}l@{}} 80.00 \\ 80.00 \end{tabular} & \begin{tabular}{@{}l@{}} 80.00 \\ 80.00 \end{tabular} & \begin{tabular}{@{}l@{}} 80.00 \\ 80.00 \end{tabular}\\
\midrule
\stt{move(A,center,C)} & \begin{tabular}{@{}l@{}} $[\vect{f_1},\vect{f_2},\vect{f_3}]$ \\ $\vect{f_1}$ as in \cite{despinoy2015unsupervised} \end{tabular} & \begin{tabular}{@{}l@{}} \textbf{40.00} \\ 22.22 \end{tabular} & \begin{tabular}{@{}l@{}} 40.00 \\ 40.00 \end{tabular} & \begin{tabular}{@{}l@{}} \textbf{40.00} \\ 28.57 \end{tabular}\\
\midrule
\stt{grasp(A,ring,C)} & \begin{tabular}{@{}l@{}} $[\vect{f_2},\vect{f_3}]$ \\ $\vect{f_1}$ as in \cite{despinoy2015unsupervised} \end{tabular} &  \begin{tabular}{@{}l@{}} \textbf{60.00} \\ 56.25 \end{tabular} & \begin{tabular}{@{}l@{}} 75.00 \\ 75.00 \end{tabular} & \begin{tabular}{@{}l@{}} \textbf{66.66} \\ 64.29 \end{tabular}\\
\midrule
\stt{extract(A,ring,C)} & \begin{tabular}{@{}l@{}} $[\vect{f_2},\vect{f_3}]$ \\ $\vect{f_1}$ as in \cite{despinoy2015unsupervised} \end{tabular} & \begin{tabular}{@{}l@{}} \textbf{100.00} \\ 12.50 \end{tabular} & \begin{tabular}{@{}l@{}} \textbf{62.50} \\ 12.50 \end{tabular} & \begin{tabular}{@{}l@{}} \textbf{76.92} \\ 12.50 \end{tabular}\\
\midrule
\stt{release(A)} & \begin{tabular}{@{}l@{}} $[\vect{f_2},\vect{f_3}]$ \\ $\vect{f_1}$ as in \cite{despinoy2015unsupervised} \end{tabular} & \begin{tabular}{@{}l@{}} 52.38 \\ 52.38 \end{tabular} & \begin{tabular}{@{}l@{}} 52.38 \\ 52.38 \end{tabular} & \begin{tabular}{@{}l@{}} 52.38 \\ 52.38 \end{tabular}\\
\midrule
\textbf{Average} & \begin{tabular}{@{}l@{}} Our method \\ $\vect{f_1}$ as in \cite{despinoy2015unsupervised} \end{tabular} & \begin{tabular}{@{}l@{}} \textbf{69.03} \\ 52.23 \end{tabular} & \begin{tabular}{@{}l@{}} \textbf{66.65} \\ 58.31 \end{tabular} & \begin{tabular}{@{}l@{}} \textbf{66.95} \\ 54.62 \end{tabular}\\
\bottomrule
\end{tabular*}
\end{table}

\subsection{Computational performance}
The computational performance of our action identification algorithm is tested on a PC using 2.6 GHz Intel Core i7-6700HQ CPU (4 cores / 8 threads). The full action identification procedure (segmentation + classification) for one execution trace requires \SI{0.45}{\s} on average (maximum \SI{0.58}{\s} for the standard execution with the highest number of actions). For changepoint filtering, the time required for fluent identification from video frames must be also accounted for. Fluent identification from a single frame requires \SI{0.09}{\s} on average; thus, fluent identification in a full execution trace requires at most \SI{2.88}{\s} for the execution with most actions (standard environmental conditions).
This results in better performance than \cite{despinoy2015unsupervised}, where \SI{5}{\s} are needed for action identification using a better CPU (3.4 GHz Intel Core i7-3770 with 4 cores / 8 threads).
In fact, one main limitation of \cite{despinoy2015unsupervised} lies in the use of persistence analysis \cite{edelsbrunner2000topological} and dynamic time warping (DTW) \cite{sakoe1978dynamic} to identify changepoints. Persistence analysis removes noise and identifies relevant local minima and maxima in the kinematic features, but still results in spurious changepoints which must be removed with further post-processing through DTW based on Euclidean distance. DTW evaluates the alignment of \emph{all possible} consecutive segments in the kinematic signature, including spurious ones. As also claimed by the autrhors of \cite{despinoy2015unsupervised}, this increases computational complexity. 
Notice that the computational performance also depends on the number of segments to be identified in execution traces.
In \cite{despinoy2015unsupervised}, 12 actions are involved in a single execution trace (with possibly some repetitions), while the highest number of segments in our execution trace under standard environmental conditions is 32; hence, the comparison of computational performance is fair.

The computational performance of our algorithm is also comparable with TSC \cite{krishnan2018transition} ($\approx1-10$ \si{\s}, though the authors do not provide hardware information and do not consider automatic extraction of visual features, presented as a future work).

\section{Conclusion}
\label{sec:conclusion}
Unsupervised segmentation of surgical actions from expert executions is crucial for robotic surgery. In this paper, we presented a novel algorithm combining kinematic and semantic visual features to identify actions in unlabeled executions of a training ring transfer task with dVRK. 
We addressed some limitations of state-of-the-art research, as identification from datasets including short actions, few executions of the task and possibly non-homogeneous (anomalous) flows of actions.
We validated our algorithm in three different experimental conditions.
We first evaluated the performance on a limited dataset of 9 executions from humans with different expertise in using dVRK. We were able to significantly improve results in the state of the art under different experimental conditions (but considering only expert executions), doubling F1-score also for short actions.

We then tested the robustness of our algorithm on a similar dataset, where a single human execution was augmented with low-frequency kinematic noise (10 executions total). Results were comparable with the previous test, though the improvement with respect to the state of the art was not similarly relevant.

Finally, we showed that our algorithm is able to improve the state of the art (especially for short actions) on a non-homogeneous dataset including only 4 executions and actions appearing rarely (only 2 times). 

The average F1-score over all actions is $58-66 \%$, comparable or better than state-of-the-art methods based on deep learning \cite{shi2021domain} ($51\%$) for a similar task to ours.
Improvements mainly rise from the inclusion of semantic visual features for classification, which compensate kinematic variability.

We also showed that our algorithm has comparable or better computational performance than state of the art, allowing feature extraction and action identification within few seconds. 

This paper represents the first step towards unsupervised learning of action-level SPMs. Next research will combine our algorithm with inductive learning of interpretable SPMs \cite{meli2020towards, meli2021inductive}, and unsupervised semantic recognition of events from raw videos \cite{gan2016concepts,sridhar2010unsupervised}.






\bibliographystyle{IEEEtran}
\bibliography{IEEEabrv,bare_jrnl}

\begin{thebibliography}{10}
\providecommand{\url}[1]{#1}
\csname url@samestyle\endcsname
\providecommand{\newblock}{\relax}
\providecommand{\bibinfo}[2]{#2}
\providecommand{\BIBentrySTDinterwordspacing}{\spaceskip=0pt\relax}
\providecommand{\BIBentryALTinterwordstretchfactor}{4}
\providecommand{\BIBentryALTinterwordspacing}{\spaceskip=\fontdimen2\font plus
\BIBentryALTinterwordstretchfactor\fontdimen3\font minus
  \fontdimen4\font\relax}
\providecommand{\BIBforeignlanguage}[2]{{%
\expandafter\ifx\csname l@#1\endcsname\relax
\typeout{** WARNING: IEEEtran.bst: No hyphenation pattern has been}%
\typeout{** loaded for the language `#1'. Using the pattern for}%
\typeout{** the default language instead.}%
\else
\language=\csname l@#1\endcsname
\fi
#2}}
\providecommand{\BIBdecl}{\relax}
\BIBdecl

\bibitem{palep2009robotic}
J.~H. Palep, ``Robotic assisted minimally invasive surgery,'' \emph{Journal of
  minimal access surgery}, vol.~5, no.~1, p.~1, 2009.

\bibitem{weksler2012robot}
B.~Weksler \emph{et~al.}, ``Robot-assisted minimally invasive esophagectomy is
  equivalent to thoracoscopic minimally invasive esophagectomy,''
  \emph{Diseases of the Esophagus}, vol.~25, no.~5, pp. 403--409, 2012.

\bibitem{daouadi2013robot}
M.~Daouadi \emph{et~al.}, ``Robot-assisted minimally invasive distal
  pancreatectomy is superior to the laparoscopic technique,'' \emph{Annals of
  surgery}, vol. 257, no.~1, pp. 128--132, 2013.

\bibitem{bharathan2013operating}
R.~Bharathan \emph{et~al.}, ``Operating room of the future,'' \emph{Best
  Practice \& Research Clinical Obstetrics \& Gynaecology}, vol.~27, no.~3, pp.
  311--322, 2013.

\bibitem{nagy2019dvrk}
T.~D. Nagy and T.~Haidegger, ``A dvrk-based framework for surgical subtask
  automation,'' \emph{Acta Polytechnica Hungarica}, pp. 61--78, 2019.

\bibitem{DeRossi2021}
G.~{De Rossi} \emph{et~al.}, ``{A First Evaluation of a Multi-Modal Learning
  System to Control Surgical Assistant Robots via Action Segmentation},''
  \emph{IEEE Transactions on Medical Robotics and Bionics}, pp. 1--11, 2021.

\bibitem{Muradore}
R.~Muradore \emph{et~al.}, ``Development of a cognitive robotic system for
  simple surgical tasks,'' \emph{Int J of Advanced Robotic Systems}, vol.~12,
  no.~4, p.~37, 2015.

\bibitem{ICAR19}
M.~Ginesi \emph{et~al.}, ``A knowledge-based framework for task automation in
  surgery,'' in \emph{2019 19th International Conference on Advanced Robotics
  (ICAR)}, Dec 2019, pp. 37--42.

\bibitem{IROS2020}
M.~Gin\vspace{0mm}esi \emph{et~al.}, ``Autonomous task planning and situation
  awareness in robotic surgery,'' in \emph{2020 IEEE International Conference
  on Intelligent Robots and Systems (IROS)}.\hskip 1em plus 0.5em minus
  0.4em\relax IEEE, 2020, pp. 3144--3150.

\bibitem{Lalys}
F.~Lalys and P.~Jannin, ``Surgical process modeling: a review,''
  \emph{International J CARS}, vol.~9, no.~3, pp. 495--511, 2014.

\bibitem{charriere2017real}
K.~Charri{\`e}re \emph{et~al.}, ``Real-time analysis of cataract surgery videos
  using statistical models,'' \emph{Multimedia Tools and Applications},
  vol.~76, no.~21, pp. 22\,473--22\,491, 2017.

\bibitem{twinanda2016endonet}
A.~P. Twinanda \emph{et~al.}, ``Endonet: a deep architecture for recognition
  tasks on laparoscopic videos,'' \emph{IEEE transactions on medical imaging},
  vol.~36, no.~1, pp. 86--97, 2016.

\bibitem{dergachyova2018knowledge}
O.~Dergachyova \emph{et~al.}, ``Knowledge transfer for surgical activity
  prediction,'' \emph{International journal of computer assisted radiology and
  surgery}, vol.~13, no.~9, pp. 1409--1417, 2018.

\bibitem{ke2013review}
S.-R. Ke \emph{et~al.}, ``A review on video-based human activity recognition,''
  \emph{Computers}, vol.~2, no.~2, pp. 88--131, 2013.

\bibitem{naftel2005classification}
A.~Naftel and S.~Khalid, ``Classification and prediction of motion trajectories
  using spatiotemporal approximations,'' in \emph{Proc. Int. Workshop on Human
  Activity Recognition and Modelling}, 2005, pp. 17--26.

\bibitem{murali2016tsc}
A.~Murali \emph{et~al.}, ``Tsc-dl: Unsupervised trajectory segmentation of
  multi-modal surgical demonstrations with deep learning,'' in \emph{2016 IEEE
  International Conference on Robotics and Automation (ICRA)}.\hskip 1em plus
  0.5em minus 0.4em\relax IEEE, 2016, pp. 4150--4157.

\bibitem{shao2018unsupervised}
Z.~Shao \emph{et~al.}, ``Unsupervised trajectory segmentation and promoting of
  multi-modal surgical demonstrations,'' in \emph{2018 IEEE/RSJ International
  Conference on Intelligent Robots and Systems (IROS)}.\hskip 1em plus 0.5em
  minus 0.4em\relax IEEE, 2018, pp. 777--782.

\bibitem{gao2014jhu}
Y.~Gao \emph{et~al.}, ``Jhu-isi gesture and skill assessment working set
  (jigsaws): A surgical activity dataset for human motion modeling,'' in
  \emph{MICCAI Workshop: M2CAI}, vol.~3, 2014, p.~3.

\bibitem{krishnan2018transition}
S.~Krishnan \emph{et~al.}, ``Transition state clustering: Unsupervised surgical
  trajectory segmentation for robot learning,'' in \emph{Robotics
  Research}.\hskip 1em plus 0.5em minus 0.4em\relax Springer, 2018, pp.
  91--110.

\bibitem{fard2016soft}
M.~J. Fard \emph{et~al.}, ``Soft boundary approach for unsupervised gesture
  segmentation in robotic-assisted surgery,'' \emph{IEEE Robotics and
  Automation Letters}, vol.~2, no.~1, pp. 171--178, 2016.

\bibitem{van2019weakly}
B.~van Amsterdam \emph{et~al.}, ``Weakly supervised recognition of surgical
  gestures,'' in \emph{2019 International Conference on Robotics and Automation
  (ICRA)}.\hskip 1em plus 0.5em minus 0.4em\relax IEEE, 2019, pp. 9565--9571.

\bibitem{FLS}
N.~J. Soper and G.~M. Fried, ``The fundamentals of laparoscopic surgery: its
  time has come.'' \emph{Bull Am Coll Surg}, vol.~93, no.~9, pp. 30--32, 2008.

\bibitem{shi2021domain}
X.~Shi \emph{et~al.}, ``Domain adaptive robotic gesture recognition with
  unsupervised kinematic-visual data alignment,'' \emph{arXiv preprint
  arXiv:2103.04075}, 2021.

\bibitem{despinoy2015unsupervised}
F.~Despinoy \emph{et~al.}, ``Unsupervised trajectory segmentation for surgical
  gesture recognition in robotic training,'' \emph{IEEE Transactions on
  Biomedical Engineering}, vol.~63, no.~6, pp. 1280--1291, 2015.

\bibitem{akdemir2008ontology}
U.~Akdemir \emph{et~al.}, ``An ontology based approach for activity recognition
  from video,'' in \emph{Proceedings of the 16th ACM international conference
  on Multimedia}, 2008, pp. 709--712.

\bibitem{roberti2020improving}
A.~{Roberti} \emph{et~al.}, ``Improving rigid 3-d calibration for robotic
  surgery,'' \emph{IEEE Transactions on Medical Robotics and Bionics}, vol.~2,
  no.~4, pp. 569--573, 2020.

\bibitem{ransac}
M.~A. Fischler and R.~C. Bolles, ``Random sample consensus: a paradigm for
  model fitting with applications to image analysis and automated
  cartography,'' \emph{Communications of the ACM}, vol.~24, no.~6, pp.
  381--395, 1981.

\bibitem{loram2006frequency}
I.~D. Loram \emph{et~al.}, ``The frequency of human, manual adjustments in
  balancing an inverted pendulum is constrained by intrinsic physiological
  factors,'' \emph{The Journal of physiology}, vol. 577, no.~1, pp. 417--432,
  2006.

\bibitem{5888646}
R.~W. {Schafer}, ``What is a savitzky-golay filter? [lecture notes],''
  \emph{IEEE Signal Processing Magazine}, vol.~28, no.~4, pp. 111--117, 2011.

\bibitem{wang1999three}
X.~Wang, ``Three-dimensional kinematic analysis of influence of hand
  orientation and joint limits on the control of arm postures and movements,''
  \emph{Biological Cybernetics}, vol.~80, no.~6, pp. 449--463, 1999.

\bibitem{pistohl2008prediction}
T.~Pistohl \emph{et~al.}, ``Prediction of arm movement trajectories from
  ecog-recordings in humans,'' \emph{Journal of neuroscience methods}, vol.
  167, no.~1, pp. 105--114, 2008.

\bibitem{truong2020selective}
C.~Truong \emph{et~al.}, ``Selective review of offline change point detection
  methods,'' \emph{Signal Processing}, vol. 167, p. 107299, 2020.

\bibitem{edelsbrunner2000topological}
H.~Edelsbrunner \emph{et~al.}, ``Topological persistence and simplification,''
  in \emph{Proceedings 41st annual symposium on foundations of computer
  science}.\hskip 1em plus 0.5em minus 0.4em\relax IEEE, 2000, pp. 454--463.

\bibitem{crowley2004context}
J.~L. Crowley \emph{et~al.}, ``Context aware vision using image-based active
  recognition,'' 2004.

\bibitem{Sch06}
S.~Schaal, ``Dynamic movement primitives-a framework for motor control in
  humans and humanoid robotics,'' in \emph{Adaptive motion of animals and
  machines}.\hskip 1em plus 0.5em minus 0.4em\relax Springer, 2006, pp.
  261--280.

\bibitem{GMCDSF19}
M.~Ginesi \emph{et~al.}, ``Dynamic movement primitives: Volumetric obstacle
  avoidance,'' in \emph{2019 19th International Conference on Advanced Robotics
  (ICAR)}, Dec 2019, pp. 234--239.

\bibitem{Ginesi2021}
M.~Gine\vspace{0mm}si \emph{et~al.}, ``Dynamic movement primitives: volumetric
  obstacle avoidance using dynamic potential functions,'' \emph{Journal of
  Intelligent and Robotic Systems}, vol. 101, no.~4, p.~20, 2021.

\bibitem{ginesi2019dmp}
M.~Ginesi, N.~Sansonetto, and P.~Fiorini, ``Overcoming some drawbacks of
  dynamic movement primitives,'' \emph{Robotics and Autonomous Systems}, p.
  103844, 2021.

\bibitem{sakoe1978dynamic}
H.~Sakoe and S.~Chiba, ``Dynamic programming algorithm optimization for spoken
  word recognition,'' \emph{IEEE transactions on acoustics, speech, and signal
  processing}, vol.~26, no.~1, pp. 43--49, 1978.

\bibitem{meli2020towards}
D.~Meli \emph{et~al.}, ``Towards inductive learning of surgical task knowledge:
  a preliminary case study of the peg transfer task,'' \emph{Procedia Computer
  Science}, vol. 176, pp. 440--449, 2020.

\bibitem{meli2021inductive}
D.~Meli, M.~Sridharan, and P.~Fiorini, ``Inductive learning of answer set
  programs for autonomous surgical task planning,'' \emph{Machine Learning},
  pp. 1--25, 2021.

\bibitem{gan2016concepts}
C.~Gan \emph{et~al.}, ``Concepts not alone: Exploring pairwise relationships
  for zero-shot video activity recognition,'' in \emph{Thirtieth AAAI
  conference on artificial intelligence}, 2016.

\bibitem{sridhar2010unsupervised}
M.~Sridhar \emph{et~al.}, ``Unsupervised learning of event classes from
  video,'' in \emph{Proceedings of the Twenty-Fourth AAAI Conference on
  Artificial Intelligence}.\hskip 1em plus 0.5em minus 0.4em\relax AAAI Press,
  2010, pp. 1631--1638.

\end{thebibliography}
\end{document}